\documentclass[sigconf]{acmart}
\usepackage{tabularx}
\usepackage{float}
\usepackage{adjustbox}

\renewcommand\footnotetextcopyrightpermission[1]{} % Removes the footnote
\acmYear{}
\acmDOI{}
\acmConference[ ]{ }{ }{ }
\acmISBN{}
\begin{document}

\title{Low Rank Factorizations are Indirect Encodings\\ for Deep Neuroevolution}

%\author{Anonymous Author(s)}
\author{Jack Garbus and Jordan Pollack}
%\orcid{TODO}
\affiliation{%
  \institution{Brandeis University}
  \city{Waltham}
  \state{Massachusetts}
  \country{USA}
}
\email{garbus@brandeis.edu}

\begin{abstract}
Deep neuroevolution is a highly scalable alternative to reinforcement learning due to its unique ability to encode network updates in a small number of bytes. Recent insights from traditional deep learning indicate high-dimensional models possess intrinsic, low-rank structure. In this work, we introduce low-rank, factorized neuroevolution--an indirect encoding through which we can search a small space of low-rank factors that enforce underlying structure across a network's weights. We compare our approach with non-factorized networks of similar and smaller size to understand how much performance can be attributed to the smaller search space. We evaluate our method on a language modeling task using transformers, as well as continuous and discrete vision-based reinforcement learning tasks. Our study shows that low-rank, factorized neuroevolution outperforms or is competitive with non-factorized neuroevolution, performing notably well on language modeling. Our results also suggest deleterious factorized mutations have a stronger negative impact on performance than deleterious non-factorized mutations, which significantly reduces the runtime on environments with early termination for bad performers. More broadly, these results show how we can use insights from backpropgation-based methods to enhance neuroevolution.

\end{abstract}

%\begin{CCSXML}
%<ccs2012>
%<concept>
%<concept_id>10010147.10010257.10010293.10010294</concept_id>
%<concept_desc>Computing methodologies~Neural networks</concept_desc>
%<concept_significance>500</concept_significance>
%</concept>
%<concept>
%<concept_id>10010147.10010257.10010293.10010309</concept_id>
%<concept_desc>Computing methodologies~Factorization methods</concept_desc>
%<concept_significance>500</concept_significance>
%</concept>
%<concept>
%<concept_id>10010147.10010257.10010293.10011809.10011812</concept_id>
%<concept_desc>Computing methodologies~Genetic algorithms</concept_desc>
%<concept_significance>500</concept_significance>
%</concept>
%</ccs2012>
%\end{CCSXML}
%
%\ccsdesc[500]{Computing methodologies~Neural networks}
%\ccsdesc[500]{Computing methodologies~Factorization methods}
%\ccsdesc[500]{Computing methodologies~Genetic algorithms}


\maketitle

\begin{acks} % acknowledgements
The authors thank Kenneth Stanley, Alexander Lalejini, and Nianwen Xue for comments on early versions of this work.
\end{acks}
\section{Introduction}\label{introduction}

Low-rank matrix factorization is a long-standing approach to uncover latent
patterns or intrinsic dimensions in data by decomposing a large matrix $m\times n$ into the product of two low-dimensional factors of sizes $m\times k$ and $k\times n$, where $k$ is the rank $(k \ll m, k \ll n)$. Each factor captures different aspects of the data; one matrix
represents a set of latent features which encode
key variability, while the other matrix shows how each data
point aligns with these features. This creates a compact and
interpretable representation of the original data that encodes the
underlying structure between points while filtering out noise. Some
argue that big data is inherently low rank \citep{udell_why_2019}, making
low-rank factorizations a natural choice for the compression and
interpretation of large models and datasets.

Recently, there has been much exciting work on low-rank deep learning
for approaches utilizing back-propagation. Weights for pre-trained models
can be compressed using low-rank factorizations
\citep{yu_compressing_2017}. Low-rank, high-dimensional networks can also be trained end-to-end
\citep{yang_learning_2020, kamalakara_exploring_2022}. Most
famously, large, pre-trained models can be fine-tuned using small,
low-rank adaptors (LoRA) \citep{hu_lora_2021} which reduce the parameter count
of the backward pass.  Some even claim that
training large models is a process of reducing the intrinsic
dimensionality of a model \citep{aghajanyan_intrinsic_2020}. If true, then this smaller, intrinsic,
low-dimensional space should contain many well-formed solutions from the
full, high-dimensional parameter space.

\begin{figure*}[ht]
\centering
\includegraphics[width=\textwidth]{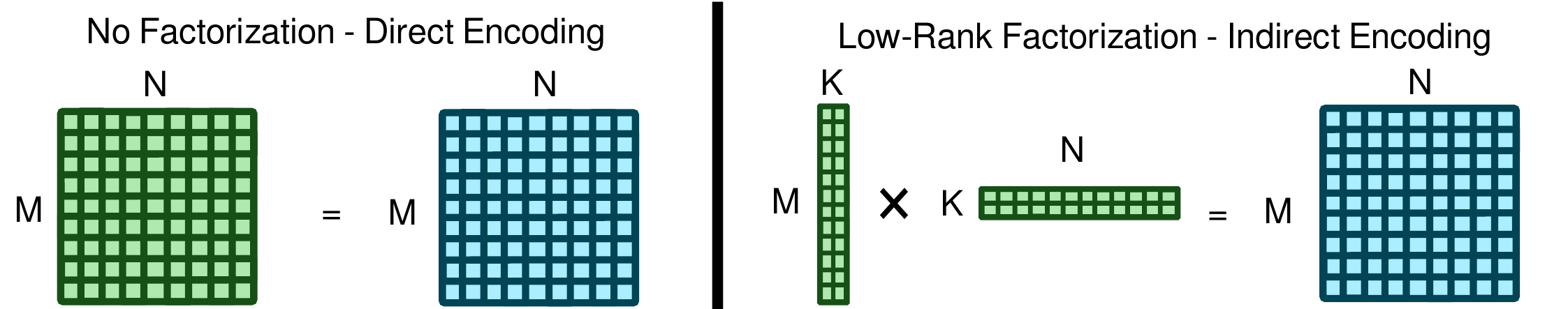}
\caption{Representations for an arbitrary weight matrix in each
representation type. Green matrices represent the genotype, i.e., the
parameters we directly mutate, whereas blue matrices represent the
phenotype, the final, developed set of parameters which are actually
used.}\label{fig:lora-vs-nofa}
\end{figure*}

This low-dimensional intrinsic space is analogous to indirect encodings used in neuroevolution—a population-based, biologically inspired approach for training neural networks through crossover and mutation, which can optimize a network's architecture and parameters. Indirect encodings--which use compact rules or patterns to represent neural network structures rather than explicitly encoding each connection--effectively reduce the search space, enforce structural regularities across the network, and significantly reduce memory requirements \citep{stanley_hypercube-based_2009, templier_geometric_2021, van_steenkiste_wavelet-based_2016}. Evolved neural networks can also be compactly represented using a small collection of random seeds, eliminating the transmission overhead when sending models to worker machines for evaluation. This enables greater horizontal scalability compared to memory-intensive methods such as reinforcement learning \citep{salimans_evolution_2017, such_deep_2018, klijn_coevolutionary_2021}. 

One would prefer to perform search in a way that is as scalable as neuroevolution yet as well-informed as back-propagation. There exists an inherent trade-off between these two, however; the more information incorporated within an update, the more expensive the update is to compute and transmit to another machine \citep{lehman_safe_2018}. To achieve scalable yet efficient updates, one would need to restrict the search space such that updates containing minimal information are likely to be effective, or at least informative, on whatever environment is being evaluated. 

To this end, we propose low-rank factorizations as an indirect encoding for evolving neural networks. By restricting the search space to networks with low-rank weight matrices, we shrink the size of the search space while still incorporating low-rank solutions and maintaining compatibility with the scalable seed-based update scheme. Compared to prior indirect encodings \citep{stanley_hypercube-based_2009, templier_geometric_2021, van_steenkiste_wavelet-based_2016}, this method is also straightforward to understand, implement, and test. We evaluate our method with a genetic algorithm (GA) on a basic language modeling task using transformers, a continuous car racing task with RGB pixel observations, and four Atari games. Additionally, to understand how much performance can be attributed solely to the smaller search space, we compare our method to small, non-factorized \emph{phenotypes} with fewer parameters than the factorized \emph{genotype}.

Overall, we find low-rank, factorized neuroevolution either outperforms or is competitive with the non-factorized approaches of both similar phenotype and genotype size, indicating that performance stems both from the smaller search space and structure of the representation. We also find deleterious mutations formed by our method are more detectable, allowing us to terminate their evaluation early, saving time and compute.
More generally, this work shows how we can take insights from traditional deep learning to enhance search in neuroevolution. We publish our code at \url{https://github.com/jarbus/Jevo.jl}.

\section{Methods}\label{methods}

When we multiply two matrices, the elements of the product are not all independent, but rather posses a relatively simple structure. For example, when we multiply a $2\times 2$ matrix $A$ with a $2\times 2$ matrix $B$, we get a $2\times 2$ product $C$:
\[
\begin{bmatrix}
a_{11} & a_{12} \\
a_{21} & a_{22}
\end{bmatrix}
\cdot
\begin{bmatrix}
b_{11} & b_{12} \\
b_{21} & b_{22}
\end{bmatrix}
=
\begin{bmatrix}
c_{11} & c_{12} \\
c_{21} & c_{22}
\end{bmatrix}
\]

where:

\[
c_{11} = a_{11}b_{11} + a_{12}b_{21}, \quad
c_{12} = a_{11}b_{12} + a_{12}b_{22}
\]

\[
c_{21} = a_{21}b_{11} + a_{22}b_{21}, \quad
c_{22} = a_{21}b_{12} + a_{22}b_{22}
\]

Notice $C$ is not a collection of random numbers; each element is a simple linear combination of some elements from $A$ and $B$. Additionally, a change in an element from $A$ or $B$ affects multiple elements in $C$. If we view $A$ as a collection of latent features and $B$ as a collection of points (i.e, weighted combinations of feature dimensions), elements in $C$ ``reuse'' both the features and points across the matrix, and thus any change in a latent feature/point affects all weights in $C$ which use that information, adding a form of ``structure'' to our weight matrix.

This method can also reduce our search space. When we represent weights as a product of $m\times k$ and $k\times n$ factors, there are only $k(m+n)$ parameters, smaller than the $mn$ parameters if $k \ll m$ and $k \ll n$ (Figure~\ref{fig:lora-vs-nofa}). Parameter-efficient fine-tuning methods like LoRA \citep{hu_lora_2021} apply this method to new weight matrices which are multiplied and added to the frozen original parameters of a large language model, greatly reducing the memory cost of the backwards pass (at the cost of some computational expressivity). This search space reduction, in combination with the aforementioned structure, is the connection between low-rank factorizations and indirect encodings.

\subsection{Experimental Settings}

For each parameter matrix in the feedforward, self-attention, embedding,
and convolutional layers, we test two representations:

\begin{enumerate}
\item
  \textbf{Non-Factorized}: Each weight and bias matrix is initialized and randomly mutated directly.
\item
  \textbf{Factorized}\footnote{This method is different than the LoRA method used in \citep{hu_lora_2021}, which adds a trainable set of low-rank parameters to a frozen, non-factorized, pre-trained model. In our method, we evolve the low-rank factorization from scratch, and \emph{do not} add the product of these factors to a non-factorized matrix.}: Each weight matrix is represented by two low-rank matrices, which are multiplied together to ``develop'' the final weight matrix; the bias is still initialized as a vector. The weight matrix is mutated by adding noise to the factors, and the bias is mutated by adding noise to the bias vector.
\end{enumerate}

Low-rank factorization of parameter matrices both alters the exploration of parameter space and the size of the parameter space. To understand how much of the performance difference is due to the size of
the parameter space, we test two sizes of non-factored models: One with the same number of parameters as the factorized \emph{phenotype}, and a smaller model with at most the same number of parameters as the factorized \emph{genotype}. We refer to this smaller setting as \textbf{Non-Factorized (Small)}

All matrices except the transformer's embedding matrix use a rank of either 4 or 1, as \citep{hu_lora_2021} shows that even very small ranks are sufficient to achieve strong performance. We use a rank of 32 for the embedding matrix to provide additional representational power to the inputs of the model.

Convolutional layers have four-dimensional weights of shape
$(input\_channels, output\_channels, height, width)$, but their
multiplication is mathematically equivalent to a feedforward layer with
$input\_channels \times height \times width$ inputs with
$output\_channels$ outputs. We thus construct convolutional weight
matrices as a product of two low-rank factor matrices of shapes
$(input\_channels \times height \times width, k)$ and
$(k, output\_channels)$, and reshape the product back to
four-dimensions. All other weight matrices are two-dimensional, and thus
we develop them by multiplying their factors via traditional matrix multiplication.

We initialize all biases as zero. For LayerNorm layers, we initialize
the scale to 1, and fix both the layer norm scale and bias to 1 and 0
respectively throughout training. We initialize all non-zero,
non-factorized weights using Kaiming initialization
\citep{he_delving_2015}. To control for the differences in standard
deviation between factors and non-factors, we use a standard deviation
of \(\sqrt{2/c}\) for the non-factored matrix and \(\sqrt{\sqrt{2}/c}\) for
the factors, where $c$ is the number of columns. In
addition, when applying a mutation rate to factors, we instead apply the
square root of the mutation rate to each, which has a same step size as
applying the standard mutation rate to the non-factored matrix.

To evaluate the data obtained from our experiments, we employ a
suite of statistical tests that are suitable for non-parametric data to determine if there are statistically
significant differences between low-rank and full-rank neuroevolution. We utilize Kruskal-Wallis Tests when comparing more than two independent groups of non-parametric data, and Wilcoxon Rank-Sum Test for pair-wise comparisons. We measure effect size using Glass's delta.

\subsection{Scalable Random Seed Encoding}

We leverage the random seed encoding described in \citep{such_deep_2018}, where a parameter vector $\theta$ for an individual at generation $g$ can be represented by its parent's parameter vector $\theta_{g-1}$ plus some noise generated by a random seed, as seen in Equation~\ref{mutation-equation}:

\begin{equation}
\theta_g = \theta_{g-1} + \sigma \epsilon ( \tau_g )
\label{mutation-equation}
\end{equation}

Here, $\sigma$ represents our mutation rate, $\epsilon$ represents our noise generation function, and $\tau_g$ represents the random seed used to generate noise. If all workers have a copy of each parent, then we can transmit the information needed to construct any child with just the parent identifier and random seed $\tau_g$ if the mutation rate, initialization function, and architecture is fixed. In the same way, we can also continuously update the saved parents on each worker, so we never need to send the entire parameter vector or lineage of seeds. This keeps data transmission to small, constant time. This encoding can be leveraged by both factorized and non-factorized networks. While neuroevolution can also optimize the architecture of a network \citep{stanley_evolving_2002}, in this work we hold network topology constant and focus only on evolving the parameters.

\section{Experiments}\label{experiments}

To understand the general strengths and weaknesses of this indirect
encoding, we evaluate our approach on both language modeling and
reinforcement learning tasks. For language modeling, we train a
decoder-only transformer as outlined in \citep{vaswani_attention_2017}
on the TinyStories dataset \citep{eldan_tinystories_2023} using Julia's
Transformers.jl library \citep{cheng_transformersjl_2025}. While prior
work evolved the weights for attention layers
\citep{tang_neuroevolution_2020}, no published work to our knowledge has attempted to
evolve weights for the full transformer architecture, which also includes
embedding and feedforward layers. For the
reinforcement learning tasks, we test our method on the classic CarRacing task provided in
Gymnasium \citep{towers_gymnasium_2024} and a subset of Atari games from
the Arcade Learning Environment \citep{bellemare_arcade_2013}. Our
setting mostly follows \citep{such_deep_2018}, which in turn mostly
follows \citep{mnih_human-level_2015}.

In all reinforcement-learning experiments, we utilize a naive genetic
algorithm (GA) inspired by \citep{such_deep_2018}. We perform truncation
selection and select parents from the truncated individuals for
reproduction uniformly at random. Unlike prior work, we sample mutation
noise from the same function we use to initialize our weights.
We use a mutation rate of 0.01 for non-factored weights and \(\sqrt(0.01)\) for factors.

\begin{table}[!ht]
    \centering
    \begin{adjustbox}{width=\columnwidth,center}
    \begin{tabular}{|c|c|c|c|}
    \hline
        & Non-Factorized & Factorized & Small Non-Fact. \\ \hline
        Population Size & 512 & 512 & 512 \\ \hline
        Truncation Size & 16 & 16 & 16 \\ \hline
        \# of Generations & 300 & 300 & 300 \\ \hline
        Mutation Rate & 0.01 & 0.01 & 0.01 \\ \hline
        Number of Blocks & 3 & 3 & 3 \\ \hline
        Number of Heads & 4 & 4 & 4 \\ \hline
        Head Dimension & 4 & 4 & 4 \\ \hline
        Hidden Dimension & 32 & 32 & 4 \\ \hline
        Embedding Rank & - & 32 & - \\ \hline
        Q, K, V, O, FF Rank & - & 4 & - \\ \hline
        FeedForward Dimension & 128 & 128 & 16 \\ \hline
        Vocabulary Size & 2048 & 2048 & 2048 \\ \hline
        \# Genotype Parameters & 99,507 & 75,955 & 11,671 \\ \hline
        \# Phenotype Parameters & 99,507 & 99,507 & 11,671 \\ \hline
        \# Sequences & 1024 & 1024 & 1024 \\ \hline
    \end{tabular}
    \end{adjustbox}
    \caption{Transformer architecture and evolution parameters for language modeling experiments. All models use ReLU activations for the feedforward layer. }
    \label{tab:tfr-settings}
\end{table} 

\subsection{Language Modeling}\label{language-modeling}

For language modeling experiments, we train on the first 1024 sequences
from the TinyStories dataset \citep{eldan_tinystories_2023}, which are
short, simple yet nontrivial children's stories generated by GPT 3.5/4. Below is an example sequence:

\begin{quote}
One day, a little girl named Lily found a needle in her room. She knew it was difficult to play with it because it was sharp. Lily wanted to share the needle with her mom, so she could sew a button on her shirt ... After they finished, Lily thanked her mom for sharing the needle and fixing her shirt. They both felt happy because they had shared and worked together. <|endoftext|>
\end{quote}

Our training data has an average of 241 tokens per sequence, with a standard deviation of 115 tokens per sequence.
We manually
preprocess this dataset to a vocabulary of 2048 tokens using Byte Pair
Encoding \citep{gage_new_1994}. In this domain, our fitness function is
simply the negative cross-entropy loss between the predicted and actual
next token, as fitness is a function we seek to maximize, unlike
traditional loss functions, which we minimize. We evolve parameters of the full original transformer decoder architecture \citep{vaswani_attention_2017}, which, to our knowledge, no prior work has attempted.

\begin{figure}
\centering
\includegraphics[width=0.45\textwidth]{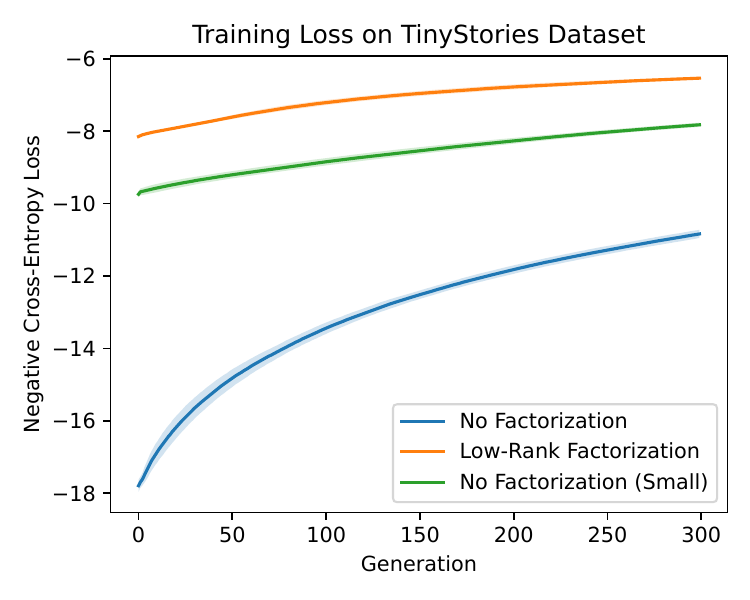}
\caption{Training loss of best individual on the first 1024 sequences of the TinyStories Dataset, averaged over 7 trials. 95\% confidence intervals are shown, but small.}\label{fig:tfr-loss}
\end{figure}

The experimental conditions for our language modeling experiments can be found in Table~\ref{tab:tfr-settings}, and our training loss curves in Figure~\ref{fig:tfr-loss}. We see both large and small non-factorized networks perform significantly worse than those with low-rank factorizations ($p \ll 0.01$, Wilcoxon; Glass's $\delta=30.4$ large, $17.6$ small). Notably, it takes hundreds of generations for small non-factorized networks to match the performance of first-generation low-rank networks, and the large non-factorized networks never even get close within our compute budget. This is remarkable, given the simplicity of our method.

We also see that the small, non-factorized solutions found are better than the large non-factorized solutions ($p \ll 0.01$, Wilcoxon). This is reasonable, given the greatly reduced search space. The performance of factorized networks, however, cannot be explained solely by the smaller genotypic search space; as shown in Table~\ref{tab:tfr-settings}, the search space of the low-rank genotype is around 76k parameters, whereas the search space of the small non-factorized phenotype is near 11k. Instead, this performance is likely due to the underlying weight structure that results from multiplying our factors.

Preliminary ablation experiments indicate the embedding matrix is the core driver of early performance. When we tested low-rank networks with non-factorized embedding matrices, performance degrades significantly; when we remove factorization from a different layer of the transformer, performance only differs from completely low-rank networks after hundreds of generations. This is notable, as the embedding matrix defines the initial representation of data as it flows through the network. It may be the case that, for a low-rank layer to provide the level of performance seen in Figure ~\ref{fig:tfr-loss}, the input for that weight must contain some underlying structure to exploit, which non-factorized embeddings initially lack. For now, we leave a rigorous investigation of this area to future work.

\begin{figure}[ht]
\centering
\includegraphics[width=0.45\textwidth]{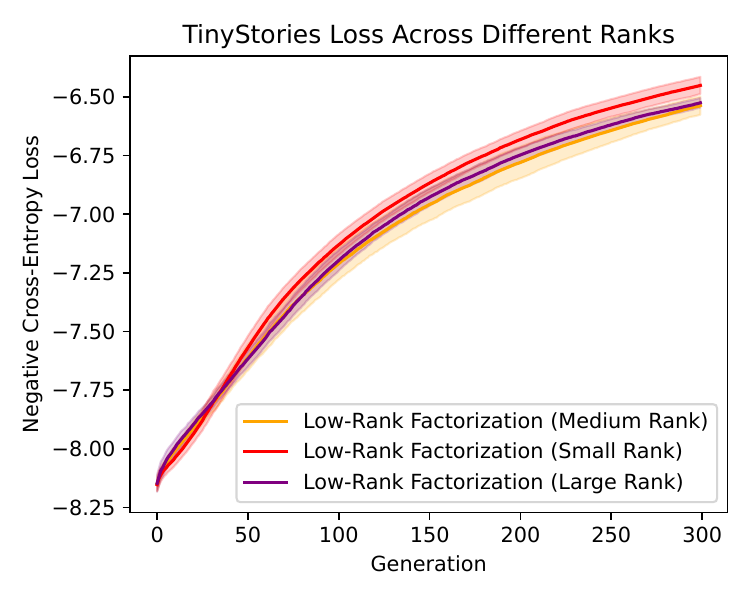}
\caption{Training loss of best individual on the first 1024 sequences of the TinyStories dataset across three ranks, averaged over 7 trials. The smallest setting has an $k_\text{embed}$=4, $k_\text{other}$=1; the medium has $k_\text{embed}$=32, $k_\text{other}$=4; large has $k_\text{embed}$=64, $k_\text{other}$=16.}\label{fig:tfr-lora-loss}
\end{figure}

To investigate how rank affects performance, we ran sets of additional factorized trials with varying rank (but identical phenotype size), shown separately in Figure~\ref{fig:tfr-lora-loss} for readability. Performance is similar between the medium and large rank ($p >0.5$, Wilcoxon, generation 300) but higher for the smallest rank ($p < 0.02$, Wilcoxon; Glass's $\delta > 1.7$ against both medium and large, generation 300). Compared to the non-factorized trials, however, the three ranks perform almost identically, further supporting the hypothesis that the improved representation--not the reduced search space--is the primary driver of performance, at least initially; rank appears plays a larger role later in evolution, when representations can benefit from further refinement. 

The results we report here are not yet competitive with the back-propagation-based results reported in \citep{eldan_tinystories_2023}, whose smallest models produce an evaluation loss of 2.38. Rather, this work is best viewed as a step towards bridging the gap with back-propagation while maintaining scalability. For research purposes, we keep our algorithm simple; we employ truncation selection and uniform reproduction. We suspect, however, that much performance remains to be captured with more powerful selection and reproduction methods.

\subsection{Reinforcement Learning}\label{reinforcement-learning}

For our reinforcement learning tasks, we performed exploratory experiments on the CarRacing task provided in Gymnasium \citep{towers_gymnasium_2024} and further experiments across a subset of Atari games \citep{bellemare_arcade_2013} which showed prior success with a GA \citep{such_deep_2018}. We use Atari settings from \citep{such_deep_2018}, which in turn follow \citep{mnih_human-level_2015}: episodes are started with up to 30 random no-op actions, we skip every 4 steps but stack the previous 4 frames. We terminate episodes after 10,000 frames, or 2,500 steps. We provide full experimental parameters in Table~\ref{rl-settings}.

\subsubsection{CarRacing}

\begin{figure}[h]
    \centering
    \includegraphics[width=0.75\linewidth]{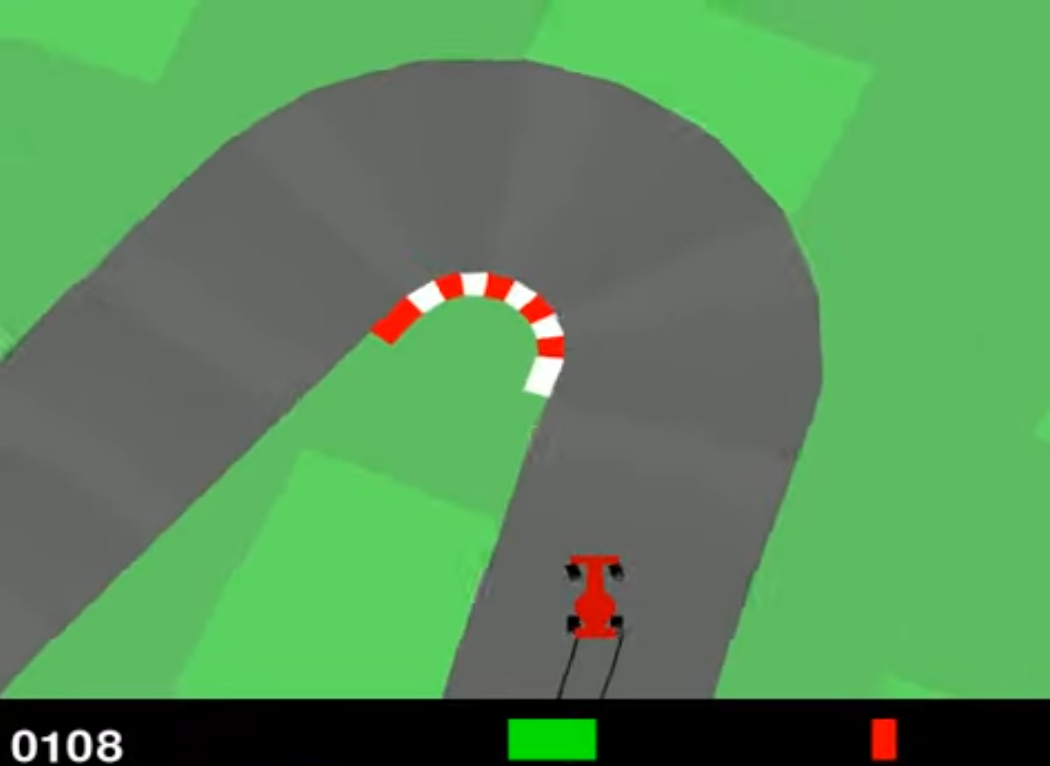}
    \caption{CarRacing Environment. Agents must learn to quickly drive a lap around a randomly generated track, given $96\times96\times3$ RGB observations, which we resize to $64\times64\times3$.}
    \label{fig:car-racing}
\end{figure}

\begin{figure}[h]
\centering
\includegraphics[width=0.45\textwidth]{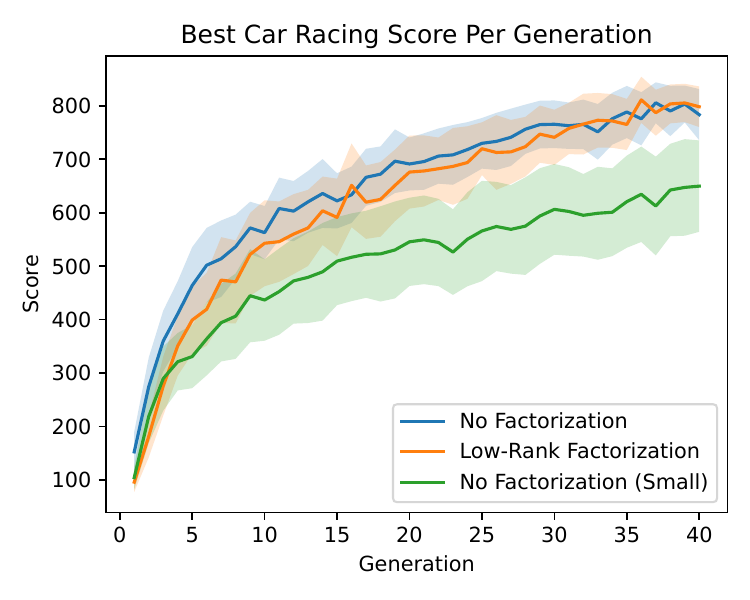} \\
\caption{Score of best individual per generation on CarRacing environment over fifteen experiments, 95\% confidence intervals are shown.}
\label{fig:car-racing-perf}
\end{figure}

CarRacing is an environment where the player drives a car around a winding track, receiving points for each unique tile of track visited (Figure~\ref{fig:car-racing}). The game ends once a player visits 95\% of the tiles on the track. We resize the original $96\times96\times3$ pixel observations down to $64\times64\times3$ and stack the last four frames as input, yielding an observation space of $64\times64\times12$. Following \citep{risi_deep_2019}, we also save compute by terminating episodes when agents go twenty steps without reaching a new tile, which occurs when agents drive off of the track. All trials are run on the CPU of the same 32-core workstation, which allows us to fairly compare the runtime of our approaches.

Tracks in CarRacing are randomly generated, so we evaluate individuals on multiple tracks to better gauge their fitness. As shown in Table~\ref{rl-settings}, for each generation of CarRacing, we evaluate all individuals two times before truncating the population to the top 32 individuals. Then, we evaluate each remaining member on four additional tracks and truncate to the top eight, before reproducing uniformly at random. We refer to these evaluation steps as ``stages''.

\begin{figure}[!h]
\centering
\includegraphics[width=0.45\textwidth]{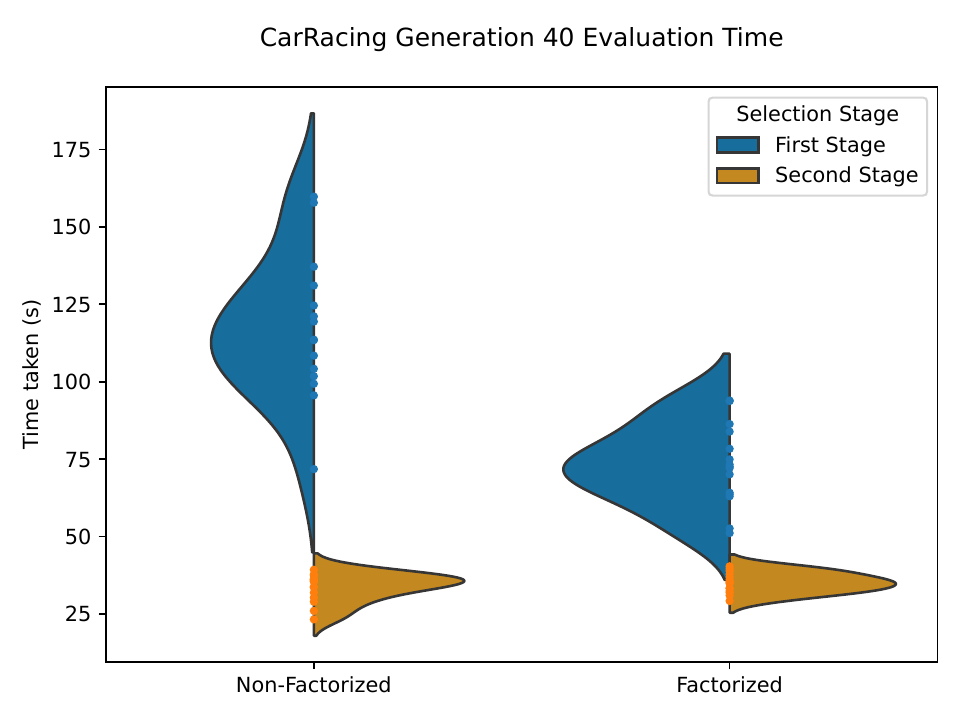}
\includegraphics[width=0.45\textwidth]{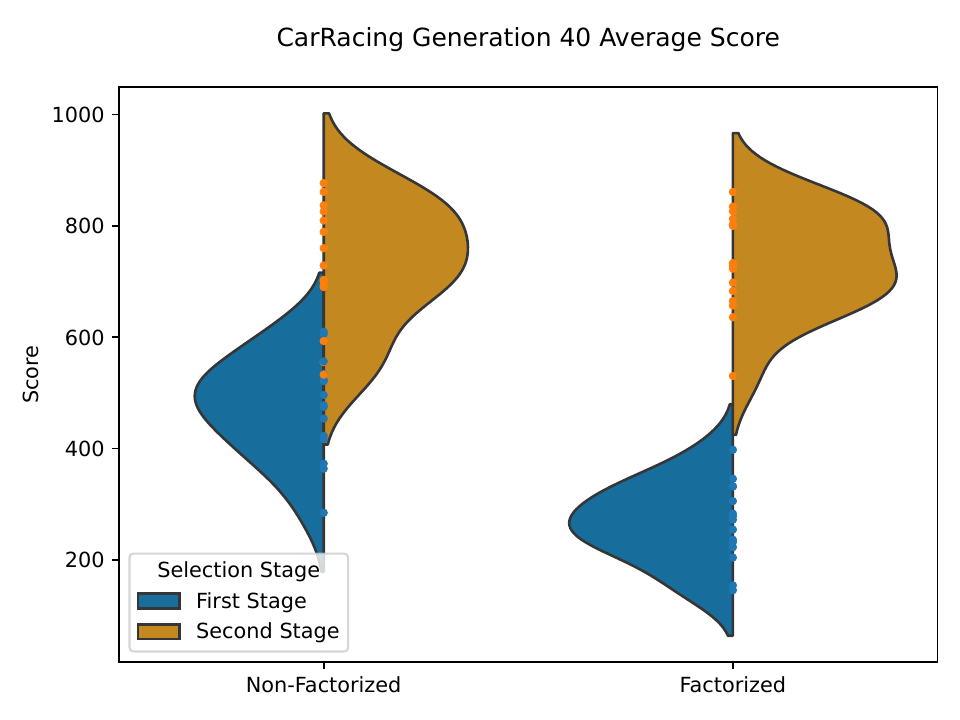} \\
\caption{Top: Violin plots of generation 40 evaluation times for each stage. Bottom: Distribution of generation 40 evaluation scores for each stage. First stage scores are the average of two evaluations for the entire population; second stage scores are the average of four evaluations for the top performers of the first stage. Our factorization method takes significantly less time ($p<0.0001$, Wilcoxon test; Glass's $\delta=-1.98$) because deleterious mutations have greater performance impact and thus fail faster. Results computed over ten trials.}
\label{fig:car-racing-last-gen}
\end{figure}

\begin{figure*}[ht]
\centering
\begin{tabular}{ccc}
\includegraphics[width=0.4\textwidth]{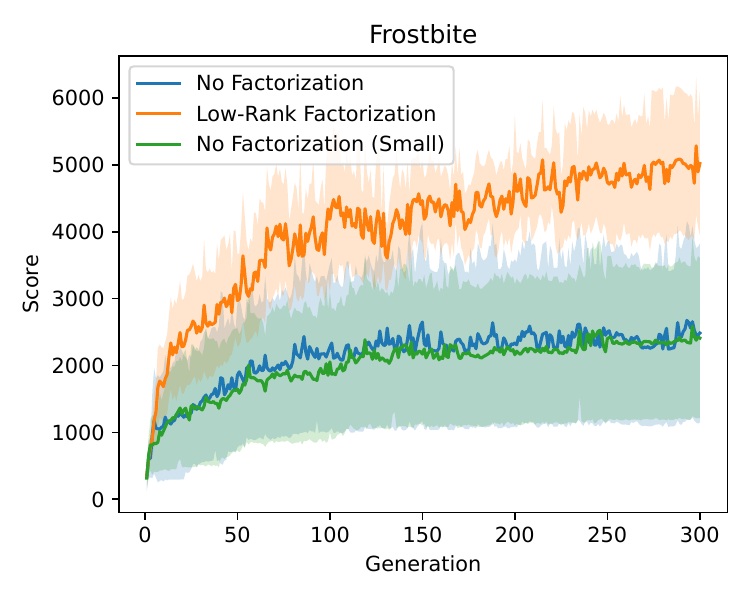} &
\includegraphics[width=0.4\textwidth]{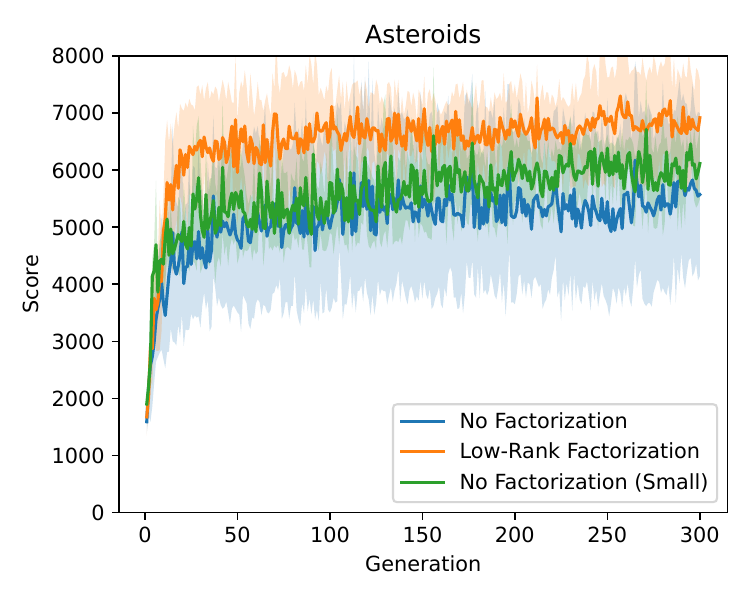} \\[1em]
\includegraphics[width=0.4\textwidth]{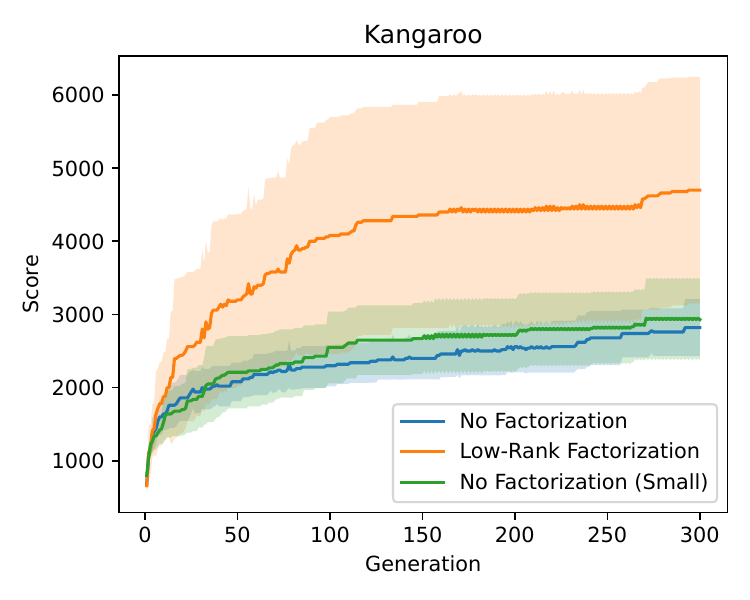} &
\includegraphics[width=0.4\textwidth]{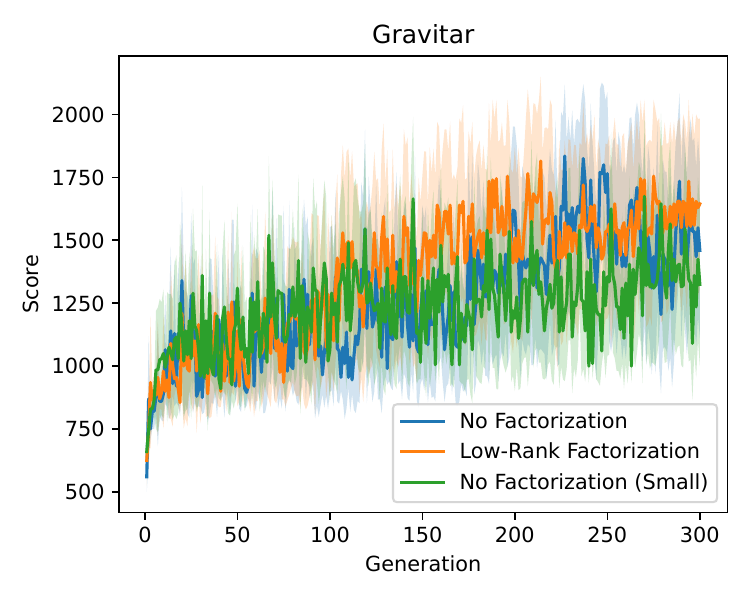}
\end{tabular}
\caption{Performance on Atari games (Frostbite, Asteroids, Kangaroo, and Gravitar) over 300 generations across ten trials; 95\% confidence intervals are shown. Y-axes are adjusted for each game.}
\label{fig:atari-scores}
\end{figure*}

Figure~\ref{fig:car-racing-perf} shows CarRacing scores of the best parent per generation across 15 experiments, averaged over six evaluations. We see that, even when restricted to a rank of one, the factorized approach still performs as well as the non-factorized approach of similar size ($p>0.8$, Wilcoxon; Glass's $\delta=0.162$), which indicates search through factorized space can consistently converge on valid solutions in our continuous RL task, and that our adjustments to factor mutation rates and standard deviations work as expected. The smaller non-factorized approach, however, performs significantly worse ($p<0.01$, Wilcoxon test; Glass's $\delta=0.91$) and displays a greater degree of variance across trials, even though it has a similar number of parameters in the genotype. The data indicates the smaller condition \emph{can} reach the high-performing solutions like the other methods, but gets stuck in local minima more frequently, unlike the factorized representation.

We noticed that our low-rank runs finished significantly faster than our non-factorized runs. The main source of this time delta is tied to the first stage of selection (Figure ~\ref{fig:car-racing-last-gen}); on generation 40, the first evaluation stage for the small, non-factorized method takes approximately 117 seconds on average, whereas the factorized method only takes approximately 72 ($p<0.0001$, Wilcoxon test; Glass's $\delta=-1.98$). The duration of the second evaluation stage--which consists of the top 32 members--is about the same, with the factorized approach taking approximately 35 seconds while the non-factorized approach takes around 33 ($p>0.5$, Wilcoxon; Glass's $\delta=0.37$).

When we analyze the performance distribution during the first stage, the cause of this discrepancy becomes clear; nonbeneficial factorized mutations hurt performance more than non-factorized ones. On generation 40, randomly mutated parents from the previous generation average a score of around 474 in the non-factorized setting and around 266 when factorized ($p<0.0001$, Wilcoxon; Glass's $\delta=-2.36$). On environments where poor performers terminate early, this is desirable; a reduction in the resources required to evaluate poor solutions frees up compute for other aspects of search, like population size or generation count.

\begin{table*}[!ht]
    \centering
    \begin{tabularx}{\textwidth}{|c|c|c|c|c|c|c|}
    \hline
        & \multicolumn{3}{c|}{CarRacing} & \multicolumn{3}{c|}{Atari} \\ \hline
        Configuration & Non-factorized & Factorized & Small Non-fact. & Non-Factorized & Factorized & Small Non-Fact. \\ \hline
        Population Size & 256 & 256 & 256 & 512 & 512 & 512 \\ \hline
        Truncation Size & 32,8 & 32,8 & 32,8 & 16 & 16 & 16 \\ \hline
        \# of Generations & 150 & 150 & 150 & 300 & 300 & 300 \\ \hline
        Mutation Rate & 0.01 & 0.01 & 0.01 & 0.01 & 0.01 & 0.01 \\ \hline
        \# evaluations & 2,4 & 2,4 & 2,4 & 1 & 1 & 1 \\ \hline
        Max Steps per Episode & - & - & - & 2500 & 2500 & 2500 \\ \hline
        Frame Stack & 4 & 4 & 4 & 4 & 4 & 4 \\ \hline
        Frame Skip & - & - & - & 4 & 4 & 4 \\ \hline
        Conv1 & 12$\rightarrow$32, 4x4, 2 & 12$\rightarrow$32, 4x4, 2 & 12$\rightarrow$4, 4x4, 2 & 12$\rightarrow$32, 8x8, 4 & 12$\rightarrow$32, 8x8, 4 & 12$\rightarrow$4, 8x8, 4 \\ \hline
        Conv2  & 32$\rightarrow$64, 4x4, 2 & 32$\rightarrow$64, 4x4, 2 & 4$\rightarrow$4, 4x4, 2 & 32$\rightarrow$64, 4x4, 2 & 32$\rightarrow$64, 4x4, 2 & 4$\rightarrow$8, 4x4, 2 \\ \hline
        Conv3  & 64$\rightarrow$128, 4x4, 2 & 64$\rightarrow$128, 4x4, 2 & 4$\rightarrow$8, 4x4, 2 & 64$\rightarrow$64, 3x3, 1 & 64$\rightarrow$64, 3x3, 1 & 8$\rightarrow$8, 3x3, 1 \\ \hline
        Conv4  & 128$\rightarrow$256, 4x4, 2 & 128$\rightarrow$256, 4x4, 2 & 8$\rightarrow$16, 4x4, 2 & - & - & - \\ \hline
        Dense1 & 1024$\rightarrow$256 & 1024$\rightarrow$256 & 64$\rightarrow$32 & 3136$\rightarrow$256 & 3136$\rightarrow$256 & 392$\rightarrow$32 \\ \hline
        Dense2 & 256$\rightarrow$3 & 256$\rightarrow$3 & 32$\rightarrow$3 & 256$\rightarrow$18 & 256$\rightarrow$18 & 32$\rightarrow$18 \\ \hline
        Rank (each layer) & - & 1 & - & - & 4 & - \\ \hline
        Genotype \# Parameters & 957,923 & 6,980 & 5,795 & 902,066 & 26,674 & 17,350 \\ \hline
        Phenotype \# Parameters & 957,923 & 957,923 & 5,795 & 902,066 & 902,066 & 17,350 \\ \hline
    \end{tabularx}
    \caption{Experiment parameters for CarRacing and Atari models. Convolutional cells are in the form (input channels$\rightarrow$ output channels, kernel, stride). In CarRacing, tracks are randomly generated, so we perform evaluation in two stages to determine the true elite, in a manner similar to \citep{risi_deep_2019}. We evaluate all individuals two times, truncate the population to the top 32 members, then evaluate each remaining member another four times and truncate to the best eight individuals. All models use ReLU activations, and each layer in our factorized model uses the same rank, except for the output layer.}
    \label{rl-settings}
\end{table*}

\subsubsection{Atari}

We evaluate each method across ten trials on four atari games where GAs have showed prior success \cite{such_deep_2018}: Asteroids, Frostbite, Kangaroo, and Gravitar; preliminary experiments indicate our method does not help on games where GAs greatly struggle. We resize the original $210\times160\times3$ RGB observations to $84\times84\times3$, skip every 4 frames, set the probability of repeating actions to zero, and select actions using argmax. To make Atari games stochastic, like \citep{such_deep_2018}, we perform up to 30 random no-op actions at the start of the evaluation before handing control to the agent for the remainder. As the Atari experiments only perform one evaluation stage per generation, run on various cluster nodes with different hardware, and exhibit greater performance differences between methods, we cannot run a fair runtime and score comparison as done for CarRacing in Figure~\ref{fig:car-racing-last-gen}.

Compared to CarRacing, our Atari results are noisy (Figure~\ref{fig:atari-scores}), likely due to the greater domain complexity, variance of running one evaluation per individual, and reward sparsity. We observe, however, that factorized networks still outperform both sizes of non-factorized networks on Frostbite ($p<0.012$, Kruskal-Wallis; $p<0.016$, Wilcoxon; Glass's $\delta=1.31$). 

While our method appears to outperform the baselines on Asteroids and Kangaroo in Figure~\ref{fig:atari-scores}, ten trials proves insufficient to claim statistical significance at generation 300 due to the extreme variance of these experiments (Asteroids: $p\approx0.16$, Kangaroo: $p\approx0.10$; Kruskal-Wallis). However, our results on Kangaroo are still notable; the best trial across both non-factorized methods reaches 4,400, while three factorization trials find solutions with scores 8,800, 8,600, and 7000 early on. This further supports the notion that performant solutions have low-rank structure which our representations can exploit.

Crucially, there is \emph{no} Atari game where factorized neuroevolution hurts performance--only on Gravitar is performance clearly similar to the baselines ($p>0.4$, Kruskal-Wallis). Altogether, despite the variance, these results hold immense promise, and clearly indicate that search in low-rank, factorized space is as effective as, if not superior to, non-factorized neuroevolution. 

\section{Discussion}\label{discussion}

While our method achieves promising performance, it cannot explore the full space of its parameters like a conventional method can. On some problems, we can imagine that a solution may be well approximated by low-rank factorization but still require a full-rank weight matrix to achieve optimal performance. On the language modeling task, we found we could add a low-rank product and a non-factorized weight matrix (initialized to zero) together to combine the best of both worlds--like LoRA \citep{hu_lora_2021}, but reversed. In these experiments, neuroevolution quickly honed in on low-rank solutions before refining them using the more expressive non-factorized matrix. There are doubtless a plethora of possible approaches to search across various underlying dimensions in parallel.

It is also interesting how large networks of low-rank complete a generation of CarRacing 38\% faster than non-factorized networks. This demonstrates how scale is not only tied only to operations per second or elite performance. Rather, on environments with early termination, scale is also tied to how quickly we can dismiss non-beneficial solutions, particularly those which are just slightly worse than their parent, as these require more compute to distinguish. We believe this does not hurt the diversity of search, but rather enhances it, as low-rank search explores solutions with greater differences in behavior and less insignificant mutations.

Surprisingly, factorization \emph{never} hurts performance, even with a rank of 1. This is notable, as this method is no minor tweak--we are fundamentally altering the search space and exploration trajectory of evolution. This reinforces the idea that, across language modeling and reinforcement learning tasks, solutions are inherently low-rank, and there is little doubt that more sophisticated tensor decomposition methods can push this improvement much further. 

Our objective here is not to claim state-of-the-art performance on a benchmark, but rather to demonstrate how we can use insights from traditional deep learning to bias evolutionary search towards solutions similar to those found via back-propagation. To this end, there are no shortage of future directions, thanks to significant progress in areas such as parameter-efficient fine-tuning, quantization, and interpretability \citep{liu_dora_2024,frantar_gptq_2023, templeton_scaling_2024}.

Perhaps the most interesting direction for future work lies in the combination of pre-trained models and well-formed updates, as factorization significantly outperforms baselines on language modeling. Methods like low-rank factorization may enable neuroevolution to effectively leverage pre-trained embeddings, act as a scalable and gradient-free post-training step for language models, or even power coevolutionary arms races in agentic settings.

As said in \emph{The Bitter Lesson}: ``The biggest lesson that can be read from 70 years of AI research is that general methods that leverage computation are ultimately the most effective, and by a large margin''\citep{sutton_bitter_2019}. Gradient-based methods work well on single machines and small clusters, but significant engineering challenges arise at scale, particularly around networking and power, as nodes must be physically located near each other to use high-bandwidth cables \citep{patel_multi-datacenter_2024}. As hardware, software, and manufacturing improvements decrease the cost of inference alongside the rise of reasoning models \citep{deepseek-ai_deepseek-r1_2025}, neuroevolution can easily scale to leverage not only the latest, but also cheapest compute in the world using compact, indirect encodings. 

\section{Conclusion}\label{conclusion}

Our work introduces low-rank factorizations as an indirect encoding for neuroevolution. We show that low-rank factorization reduces the search space and enforces a performant underlying weight structure while maintaining compatibility with highly-scalable seed-based encoding schemes. We incorporate our encoding method into a genetic algorithm and show that on a basic language modeling task, continuous car racing task, and a subset of Atari games, this indirect encoding is competitive with or outperforms our baselines.
We also find that mutations in factorized space have greater impact on performance, which reduces runtime for environments that terminate poor performers early during evaluation. More generally, this work shows how we can take artifacts of back-propagation and apply them to neuroevolution at scale.

\bibliographystyle{ACM-Reference-Format}
\bibliography{main}
%\appendix

\end{document}